\pdfoutput=1
\documentclass[sigconf]{acmart}

\usepackage{bm}
\usepackage{csquotes}
\usepackage{cleveref}
\usepackage{balance}

\copyrightyear{2022} 
\acmYear{2022} 
\setcopyright{acmlicensed}\acmConference[AIES'22]{Proceedings of the 2022 AAAI/ACM Conference on AI, Ethics, and Society}{August 1--3, 2022}{Oxford, United Kingdom}
\acmBooktitle{Proceedings of the 2022 AAAI/ACM Conference on AI, Ethics, and Society (AIES'22), August 1--3, 2022, Oxford, United Kingdom}
\acmPrice{15.00}
\acmDOI{10.1145/3514094.3534154}
\acmISBN{978-1-4503-9247-1/22/08}

\begin{document}

\fancyhead{}

\title{Equalizing Credit Opportunity in Algorithms: Aligning Algorithmic Fairness Research with U.S. Fair Lending Regulation}

\author{I. Elizabeth Kumar}
\affiliation{
  \institution{Brown University}
  \country{USA}
  }
\email{iekumar@brown.edu}

\author{Keegan E. Hines}
\affiliation{
  \institution{Arthur AI}
  \country{USA}
  }
\email{keegan@arthur.ai}

\author{John P. Dickerson}
\affiliation{
  \institution{Arthur AI}
  \country{USA}
  }
\email{john@arthur.ai}

\begin{abstract}
Credit is an essential component of financial wellbeing in America, and unequal access to it is a large factor in the economic disparities between demographic groups that exist today. Today, machine learning algorithms, sometimes trained on alternative data, are increasingly being used to determine access to credit, yet research has shown that machine learning can encode many different versions of “unfairness,” thus raising the concern that banks and other financial institutions could---potentially unwittingly---engage in illegal discrimination through the use of this technology. In the US, there are laws in place to make sure discrimination does not happen in lending and agencies charged with enforcing them.
However, conversations around fair credit models in computer science and in policy are often misaligned: fair machine learning research often lacks legal and practical considerations specific to existing fair lending policy, and regulators have yet to issue new guidance on how, if at all, credit risk models should be utilizing practices and techniques from the research community. This paper aims to better align these sides of the conversation.
We describe the current state of credit discrimination regulation in the United States, contextualize results from fair ML research to identify the specific fairness concerns raised by the use of machine learning in lending, and discuss regulatory opportunities to address these concerns.
\end{abstract}




\maketitle

\section*{Introduction}
\addcontentsline{toc}{section}{Introduction} 

Credit is an essential component of financial well-being for Americans, and unequal access to it is a significant factor in the economic disparities between demographic groups that exist today. For this reason, it is critical to make sure the American lending ecosystem is free of discrimination. In America, there are laws in place which specifically ban discrimination in lending, as well as agencies charged with enforcing them. Today, machine learning (ML) algorithms (sometimes trained on ``nontraditional'' data) are increasingly being used to allocate access to credit. A vast body of research has demonstrated that ML algorithms can encode many different versions of ``unfairness,'' thus raising the concern that banks and other financial institutions could—potentially unwittingly--engage in illegal discrimination through the use of this technology.

The nebulous threat of ``algorithmic discrimination'' poses a challenge to federal regulators, who must decide how, if at all, to update their enforcement practices or issue new guidance in light of these concerns \cite{interagency_rfi}, which are often articulated by computer scientists in the abstract and not in terms of the actual practices, data, and algorithms used in this sector.
Meanwhile, without specific guidance from regulators, researchers and practitioners who \emph{want} to study or apply fair ML in this particular setting lack a clear picture of the kinds of tools and metrics that will be useful, legal, and practical for detecting and correcting unfairness in algorithms in this setting.
For these reasons, this paper aims to orient the conversation around fair ML research in the context of predicting credit risk from \emph{both} perspectives.

In~\Cref{sec:prelims}, we briefly describe the state of American fair lending regulation and analyze recent messaging from certain federal agencies on the threat of algorithmic discrimination.
In~\Cref{sec:fairml}, we discuss methods proposed by the ML community to measure unfairness in algorithms, and determine the extent to which they may relate to the principles of the Equal Credit Opportunity Act (ECOA) and the goals of the federal agencies discussed above. Keeping these metrics in mind, we contextualize results from fair ML research in the consumer credit setting, and identify specific fair lending risks throughout different parts of a machine learning system's development. By analyzing how these mechanisms are likely to play out in the credit setting, we can be more specific about the kinds of problems regulators should anticipate and address, rather than repeating the folk wisdom of "bias in, bias out."
Finally, in \Cref{sec:regulatory-opps}, we discuss specific opportunities for regulators to use their authority to encourage fair ML practices.

\section{Credit discrimination regulation in the United States}\label{sec:prelims}

In this section, we provide a background on the laws and policies which regulate anti-discrimination in consumer credit. We further set the stage for the conversation about algorithmic discrimination by identifying specific comments and actions from federal agencies signifying their willingness to tackle the issue of discrimination in algorithms.

\subsection{Fair lending legislation}

\subsubsection{Anti-discrimination legislation}

The issue of discrimination in credit lending decisions is not novel to the algorithmic setting. While lending has been around for centuries, Americans increasingly began to rely on consumer credit to finance large purchases in the 1950's and 60's \citep{ritter}. During this period, individual loan officers and specialists were ultimately responsible for the subjective determination of whether a loan applicant was creditworthy; numerical methods for estimating credit risk existed but were not widely or systematically used \citep{scored_society}. This presented a risk of intentional discrimination due to personal bias. Additionally, some codified lending policies in effect at the time clearly disadvantaged women and minorities. During congressional hearings, testimonies cited practices such as requiring single women to provide a male co-signer for a mortgage loan \citep{ritter, freeman_racism_2017}.

In the spirit of implementing ideas from the civil rights legislation of the 60's, which did not directly address lending, ECOA was passed in 1974 to ensure that all Americans were treated fairly in a system that determined so much of their economic success. It prohibits creditors from ``discriminat[ing] against any applicant, with respect to any aspect of a credit transaction on the basis of race, color, religion, national origin, sex or marital status,'' among other factors \citep{ECOA-1974}. The law applies to any organization that extends credit, including loans and credit cards.

The Fair Housing Act, also known as Title VIII of the Civil Rights Act of 1968, prohibits discrimination in housing on the basis of several protected characteristics, and applies to mortgage providers. The U.S. Department of Housing and Urban Development (HUD) enforces the Fair Housing Act, and has specified narrow rules making disparate impact litigation difficult; partly because of this, mortgage algorithms are not our main focus in this paper.

\subsubsection{Data collection rules}

At the time of its passing, the ECOA gave the Board of Governors of the Federal Reserve Board (FRB) rulemaking authority to implement the law; this set of rules is known as Regulation B. Regulation B specifically prohibits the collection of information about protected characteristics: "A creditor shall not inquire about the race, color, religion, national origin, or sex of an applicant or any other person in connection with a credit transaction" \citep{regb}. Credit transactions, here, can include things like consumer credit, business credit, mortgage loans, and refinancing.

A glaring set of exceptions to this rule are in cases where the Home Mortgage Disclosure Act (HMDA) applies. Passed in 1975, the act requires certain financial institutions to provide mortgage data to the public, and in particular requires lenders to collect and report race and gender information of mortgage applications. The act was drafted in response to the practice of redlining, in which lenders would explicitly identify geographic regions and neighborhoods that they would not lend to because they were inhabited by people of color. This information is used to identify indicators of mortgage discrimination and encourage lenders to comply with ECOA \citep{taylor}.

In the non-mortgage setting, Regulation B contains an additional exception to the ban on collecting protected characteristics: when the information is explicitly collected for self-testing, which is defined as any inquiry ``designed and used specifically to determine the extent or effectiveness of a creditor’s compliance with the Act'' \citep{regb}. In doing so, lenders must make it clear to the applicant that providing the information is voluntary. However, this practice is very uncommon; \citet{slaughter_2021} speculate that this is because of a "fear that their collection of the data will validate or exacerbate claims that their decisions are biased." Self-testing might also be disincentivized if corporations believe that the data itself would ultimately benefit plaintiffs in a potential disparate impact suit.

It may seem counter-intuitive that HMDA requires the collection of sensitive information while ECOA bans it. In fact, both HMDA's requirement for collecting sensitive information and ECOA's ban on it are controversial. Some argue that the existence of HMDA provides an important basis of evidence for lawsuits or that the policy itself caused lenders to curb their own discriminatory practices, and thus a similar provision should be in place for non-mortgage lenders \citep{taylor, abuhamad_fallacy_2019, bogen_awareness, nclc_rfi}. Others, especially banks, have argued that HMDA is unfair, costly, and leads to false accusations of illegal discrimination \cite{jay2006full}.

The Federal Reserve Board, which was responsible for enforcing ECOA until the Consumer Financial Protection Bureau (CFPB) was established, has considered removing the ban on the collection of protected information several times since the law was originally passed. In 2003, it ultimately rejected a proposal to lift the ban and mandate the collection of certain sensitive information \citep{proposal_rejection_2003}. The first reason it cited was the natural one: that creditors might use this information for discriminatory purposes; however, many members of Congress, consumer advocates and researchers found this unconvincing \citep{gao}. The second was that "many creditors would elect not to collect the data while those that did collect it would use inconsistent standards, criteria and methods. Consequently, the data would be of questionable utility because there would be no assurance of its accuracy nor would there be any way to compare it from creditor to creditor" \citep{taylor}. The U.S. Government Accountability Office found in 2008 that while such a mandatory data collection could provide benefits to researchers and regulators, it could be costly or difficult for the lenders themselves \citep{gao}; \citet{bogen_awareness} suggest that the failure to implement such measures has largely been due to pressure from banks, which \citet{taylor} found were the overwhelming dissenting voice in responses to the FRB's request for comments on their proposal.

\subsection{Fair lending in practice}\label{sec:prelims-practice}

The two major discrimination doctrines which are relevant to fair lending law today are disparate treatment and disparate impact. Disparate treatment applies when individuals are explicitly treated differently on a prohibited basis. Under disparate impact doctrine, on the other hand, a creditor may be found to have illegally discriminated against a protected class if the effect of the practice adversely impacts that group even if the policy in question was facially neutral. The Supreme Court has found that the disparate impact is cognizable under the FHA \citep{inclusive_communities}, but has not made a similar ruling about ECOA. However, the court's language in Inclusive Communities \citep{taylor_history}, relevant case law \citep{credit_discrimination}, and the CFPB's official interpretation of Regulation B \citep{official_interpretation} all support the general consensus that disparate impact theory is cognizable under ECOA. Federal courts have consistently upheld this since the 1980s \cite{nclc_rfi}.

Plaintiffs usually rely on burden-shifting systems for establishing a \textit{prima facie} claim under both theories, which can then be rebutted by the defendant. For a disparate treatment case, most circuit courts have found that a modification of the McDonnell-Douglas test, originally developed by the Supreme Court in an employment discrimination case \cite{mcdonnell}, can be applied to an ECOA claim--but there is no official nationwide rule on the issue \citep{credit_discrimination}. In a prima facie disparate impact claim \cite{disparate_impact}, a plaintiff must point to a specific policy or action taken by the defendant that had a disproportionately adverse impact on members of a protected class. The defendant may respond by arguing there is a legitimate business necessity for the policy. Then, the plaintiff can respond by arguing there was a less discriminatory alternative that the defendant refused to use.

In a disparate impact claim, expert statistical testimony is necessary to demonstrate that an adverse impact exists and is disproportionately felt by members of a protected class \citep{credit_discrimination}. Again, we lack official Supreme Court guidance on how exactly to go about this under ECOA. In employment discrimination cases, however, the ratio of the proportion of protected class that receives a favorable outcome and the proportion of the control class is used; the oft-cited "80\% rule" is related to this statistic \citep{uniform_guidelines}. A related metric with precedence in the credit setting is the standardized mean difference of outcomes for two groups \citep{hall_us_fair_lending}. However, it is usually insufficient to simply compare the approval rates of two groups of applicants; since information related to creditworthiness is generally available, higher courts generally expect that plaintiffs will compare the selection rates of \textit{qualified} applicants \citep{credit_discrimination}. For this reason, statistical evidence which controls for drivers of creditworthiness--such as a conditional marginal effects test--are seen as more appropriate by federal agencies \citep{cfpb_2015}.

It is difficult for plaintiffs to find evidence that an individual lending decision was discriminatory, especially in the non-mortgage setting where sensitive attribute data about applicants is generally unavailable; some lower courts have historically acknowledged this \citep{taylor}. \citet{bogen_awareness} point out that "one of the few, robust public studies on credit scores and discrimination in the United States was performed by the FRB in 2007, at the direction of Congress. To conduct its analysis, the FRB created a database that, for the first time, combined sensitive attribute data collected by the Social Security Administration with a large, nationally representative sample of individuals’ credit records... this unusual undertaking would not have been possible without significant governmental time and resources."
Interestingly, the CFPB has worked around this data limitation in some of its enforcement actions by imputing racial information using Bayesian Improved Surname Geocoding (BISG) to amass evidence of disparate impact \cite{brookings}. On the other hand, in the mortgage setting where data is available, HMDA data alone cannot prove or disprove discrimination, and the results of discrimination studies using HMDA data are usually contentious  \cite{pager}.

An important precedent is, of course, the general acceptance of traditional credit scores as a basis of loan underwriting. Like the machine learning algorithms which are the focus of this paper, credit scores are functions of data which are meant to provide a quantitative basis on which to make a lending decision.
As of yet, there have not been successful challenges against credit scores using disparate impact theory \cite{hurley_credit_2017}. A combination of factors has contributed to this, but one seems to be that official CFPB interpretations of ECOA and OCC guidance on models are fairly generous as to what counts as a business necessity and relation to creditworthiness, respectively \cite{credit_discrimination}. Further complicating this matter is the fact that creditors tend to (credibly) argue that their scoring methods tend to \textit{expand} credit to minority applicants when compared to other methods. The FRB bolstered the credit score's ubiquity in their analysis of the 2007 database: they claimed that while credit scores have a ``differential effect" \cite{frb_study_07}, they did not "produce a disparate impact" \cite{frb_study} because credit characteristics do not act as "proxies for race and ethnicity" according to their own definition (which we will discuss the limitations of in Section 2).

\subsection{Agency communications on fair lending in algorithms}\label{sec:prelims-comms}

In this section, we analyze recent messaging from several federal agencies on the threat of algorithmic fairness in finance and credit. These agencies are generally allowed to operate independently, but many have been known to act cooperatively and take a unified stance on the interpretation of the law \citep{cfpb_empirical}. The OCC, FRB, FDIC, and CFPB recently issued a rare joint request for information regarding the use of artificial intelligence (AI) in financial services, inquiring, among other things, whether banks and other interested parties feel that additional regulatory guidance on the matter is necessary \cite{interagency_rfi}. Their response to the threat of algorithmic discrimination will be highly influential since, as Alex Engler has argued, "major legislative changes to AI oversight seem unlikely in the near future, which means that regulatory interventions will set precedent for the government’s approach to protecting citizens from AI harms" \citep{engler_6_developments}.

\subsubsection{The Consumer Financial Protection Bureau (CFPB)}


The CFPB was created by the Dodd-Frank Wall Street Reform and Consumer Protection Act in 2011. It was designed to consolidate responsibilities from several other agencies such as the Federal Reserve, FTC, and FDIC, to write and enforce rules for both bank and non-bank financial institutions. It has situated itself as being well-posed to tackle new regulatory challenges introduced by technology. The CFPB's internal ``Office of Competition and Innovation," dedicated in part to addressing these challenges, has taken initiatives such as holding tech sprints, issuing no-action letters (NALs), and developing compliance assistance sandboxes.

The most relevant action the CFPB has taken with respect to algorithmic discrimination was its NAL to fintech lending company Upstart in 2017. Upstart provided detailed public (and some private) information about its underwriting process with the bureau and requested a formal statement from the CFPB that they would not trigger any enforcement action \cite{NAL_request}. The CFPB granted the NAL. Part of the terms of the letter stipulated that Upstart would send the CFPB updates "regarding the loan applications it receives, how it decides which loans to approve, and how it will mitigate risk to consumers, as well as information on how its model expands access to credit for traditionally underserved populations" to "further its understanding of how these types of practices impact access to credit generally and for traditionally underserved populations, as well as the application of compliance management systems for these emerging practices" \cite{cfpb_NAL}.

The CFPB has been criticized for this move because of Upstart's usage of educational data in its algorithm. Several Democratic senators wrote a letter to the CFPB arguing that using this information could result in discrimination against minorities, and further argued that NALs should not be issued to provide immunity from ECOA in general \citep{harris_warren}. A group of advocacy organizations expressed concern that the Upstart NAL was issued without "fully accounting for certain aspects of the company’s model that have long been recognized as having a disparate impact on borrowers of color" and pointed out that the CFPB did not attempt to replicate the company's fair lending analysis \citep{aclu_rfi}. 

Under the Biden administration, the CFPB is expected to enforce fair lending laws more aggressively than it did under Trump. Biden "has pinpointed the agency as a key weapon in his arsenal to address racial disparities in access to loans, capital and credit" \citep{nbc_cop}. To do this, the CFPB is likely to utilize disparate impact doctrine as it did under Obama, even in cases "where disparate racial impact was unintentional" \citep{kslaw}. This proactive regulatory behavior will be partly aimed directly at the algorithmic discrimination issue. Biden's CFPB Director, Rohit Chopra, has repeatedly remarked that the agency will "closely watch for digital redlining, disguised through so-called neutral algorithms, that may reinforce the biases that have long existed" \citep{cfpb_digital_redlining}.




\subsubsection{The Federal Trade Commission (FTC)}

The FTC is tasked with protecting consumers in the United States, and thus shares the power to enforce ECOA with the CFPB. In particular, they are responsible for the regulation of non-bank financial service providers.

The FTC has positioned itself as particularly concerned with algorithmic discrimination.
In 2016, it acknowledged the potential of alternative credit scores to help expand credit to populations previously deemed unscorable, such as consumers without a credit history but that nonetheless pay their rent on time or own a car; it also pointed out that algorithmic credit decisions with a disparate impact on a protected class could violate ECOA, noting that it has taken enforcement action using this doctrine in a mortgage case \cite{ftc_bigdata}.
A 2021 blog post by staff at the FTC's Bureau of Consumer Protection posited that "apparently neutral technology can produce troubling outcomes – including discrimination by race or other legally protected classes" and indicated that the FTC Act's prohibition of unfair or deceptive practices would include the sale or use of racially biased algorithm \cite{ftc_blog}.

Most recently, Commissioner Rebecca Slaughter published a report on "Algorithms and Economic Justice" \citep{slaughter_2021}, stating that ECOA "can and should be aggressively applied" to threats of algorithmic discrimination. Notably, Slaughter expressed a personal opinion that "as with mortgage data, all other kinds of credit should be monitored by creditors consciously for disparities on the basis of protected status," and advocated for the collection of protected class data to enable firms to self-test their algorithms for fairness under Regulation B's existing exceptions.

\subsubsection{The Office of the Comptroller of the Currency (OCC)}

The OCC, which was established by the National Currency Act of 1863, is meant to regulate and charter the nation's banks. In particular, they make sure national banks and federal savings associations "operate in a safe and sound manner, provide fair access to financial services, treat customers fairly, and comply with applicable laws and regulations" \cite{occ_website}. While the OCC has traditionally only regulated traditional banks, the question of whether it should be in charge of regulating "fintech" companies has been fraught and remains legally unresolved \cite{fintech_charter}. 

The OCC notably issues guidance to banks on how to reduce risk in the development and use of mathematical models. In 2011, together with the FRB, the OCC issued Supervision and Regulation Letter 11-7 (SR 11-7), entitled Supervisory Guidance on Model Risk Management. The document describes "key aspects of an effective model risk management framework, including robust model development, implementation, and use; effective validation; and sound governance, policies, and controls" \cite{sr_11_7}. While this document does not explicitly mention illegal discrimination as a risk, in 2016, then-Comptroller Thomas J. Curry suggested that ECOA violations could be construed as such \cite{curry_2016}:

\begin{displayquote}
New companies and companies deploying new technology should understand and ensure their products and services comply with existing laws, such as the Equal Credit Opportunity Act ... Lenders who operate without considering these questions may be accruing underappreciated financial risks and reputational liabilities.
\end{displayquote}

\section{The risk of discrimination in credit risk algorithms}\label{sec:fairml}

While federal agencies have spoken broadly about the potential for discrimination in algorithms, this section aims to get more specific about the nature of the threat. First, we discuss different metrics for measuring fairness studied in the fair ML literature and how much they should matter from a credit discrimination perspective. Then, we use results from fair ML to point out some specific ways in which credit risk modeling is likely to induce problems with respect to those metrics throughout the development pipeline.

\subsection{Measuring unfairness in an algorithmic lending context}\label{sec:fairml-defs}

Recent work in machine learning has attempted to measure and mitigate discrimination in predictive models. In this section, we analyze how several technical measures of fairness align with the principles which are implied by ECOA and the regulatory bodies which concern themselves with enforcing it. We emphasize that none of these definitions can exactly capture whether a decision does, or does not, violate anti-discrimination law. However, many of the proposed metrics are consistent with a long history of testing for discrimination \citep{hutchinson_mitchell}.
We therefore suggest they might be useful as evidence in litigation, internal auditing, or as metrics with which to optimize a fair model.
Importantly, the theoretical properties of mathematical fairness metrics which we discuss here, such as how they relate to different data and modeling conditions, can be---and have been---formally studied. These studies, which we situate in the context of credit discrimination in Section~\ref{sec:fairml-lending}, provide intuition about which practices in algorithmic lending are likely to be problematic.

Throughout this section, we will refer to the framework of \citet{impossibility}, in which true qualities of individuals--the ideal basis for the decision-making process being learned--are referred to as \textit{construct features}. In the credit setting, proposed construct features might include qualities such as trustworthiness, reliability, and financial stability. The quality that the algorithm is trying to predict is known as the \textit{construct decision}; in the credit setting, this is often described or referred to as \textit{creditworthiness}.

In a perfect world, underwriters would have access to these construct features and use them to build a model to estimate creditworthiness. In reality, it is impossible to directly measure the qualities that define a strong credit applicant. Instead, algorithm designers must use approximations of the construct features, called \textit{observed features}, as inputs. To use supervised learning methods, the algorithm designers must have access to a historical dataset of observed features and an associated historical measurement of the construct outcome for each row; this is called the \textit{observed decision} or target variable. In the credit setting, the observed version of creditworthiness could be defined as (for instance) whether or not a historical applicant ended up defaulting on a loan they were issued within a certain amount of time;  \citet{barocas2016big} have pointed out that these modeling decisions are subjective choices that must be defended when making a business necessity defense. We will see that different assumptions about how these observational processes admit different intuitions for the appropriateness of fairness methods.

\subsubsection{Fairness as blindness (with proxy removal)}

Perhaps the most lenient implementation of ECOA would be to say that any data-driven scoring system optimized to predict some business-necessity-related outcome should be presumed nondiscriminatory as long as its inputs are not protected attributes or "proxies" for them.
An oft-cited example of what is meant by a "proxy" is the use of zip codes for racial discrimination in the illegal practice of redlining. The OCC has also pointed out that a person's primary language being Spanish is a proxy for racial or ethnic groups \cite{occ_bulletin}, and differential treatment based on this feature has resulted in discrimination enforcement in the past by the CFPB \cite{cfpb_language}.

The main problem with this logic is the difficulty of defining and identifying what a problematic proxy is, especially in light of advances in machine learning. An expansive definition of a proxy might include any feature statistically related in some way to a protected class, but the history of ECOA and its enforcement (in particular the generous guidelines around business necessity) generally support the usage of features with this quality.

More narrow definitions of "proxy" often involve the \textit{relationship} between a variable's protected-class-relatedness and its predictiveness for the task at hand. "Proxy discrimination" is defined by \citet{iowa_proxy} as a special case of disparate impact, when a variable is predictive of an outcome \emph{because} it is correlated with a sensitive attribute. An example that instantiates this line of reasoning appears in the FRB's 2007 study, which posits that ``a credit characteristic that derives its predictiveness solely by functioning as a proxy for demographics would not predict performance in a model that was estimated in a demographically neutral environment, where demographics are controlled for or where the estimation sample is limited to a single demographic group" \cite{frb_study} and goes on to argue that if credit characteristics are still predictive in a demographically neutral environment they do not cause disparate impact.
Similarly, \citet{bartlett} posit that ``scoring or pricing on a proxy variable that has significant residual correlation with race or ethnicity after orthogonalizing with respect to hidden fundamental credit-risk variables is illegitimate."

These definitions unfortunately induce thorny questions about how to quantify the slippery concepts of a variable's predictiveness and protected-class-relatedness, which are again made less well-defined by advances in machine learning. The FRB study focused on linear models, where a variable's coefficient can act as a notion of importance. However,
in the case of complex and nonlinear ML models, the question of how to measure predictiveness of an individual variable is the center of a long-standing debate \cite{wei2015variable}. For instance, some have argued that a variable is important if the learned model's output is sensitive to the input in some measurable way \cite{breiman}; some argue a variable's influence should be measured in terms of how much it improves a model within a class \cite{fisher}; some argue the variable's importance with respect to every subset of the other variables is important \cite{shap}. This question cannot be answered without encoding implicit epistemic values \cite{epistemic}, and remains not well defined within the community at large, much less in the context of a business necessity defense.

As for protected-class-relatedness, the focus on identifying the "proxy-ness" of a single variable ignores the fact that several variables taken \emph{together}, especially if a complex nonlinear model is used, can be very related to protected class information even when the individual variables are not. \citet{gillis} found that by training a race prediction model on HMDA variables, they are collectively "more predictive of race than zip code."
For these reasons, Gillis concludes that ultimately, rather than focusing on the elimination of individual proxies from credit risk models, regulatory agencies should measure the fairness of a machine learning model in terms of \emph{outcomes.} The rest of the definitions mentioned in this section all at least partly judge a model by its predictions.

\subsubsection{Equality of outcomes}

As we discussed in Section \ref{sec:prelims-practice}, comparing rates of positive and negative outcomes across groups is often used as evidence of disparate impact in employment discrimination. In ML, equal outcome rates across groups is often called demographic parity. Returning to the framework of \citet{impossibility}, if we believe that each subpopulation is similar in the construct feature space, we should assert that any differences in outcomes under an algorithm are discriminatory. However, as we also discussed in Section \ref{sec:prelims-practice}, credit scores are rarely challenged in courts even though they differ across groups due to a desire to compare outcome rates among \textit{qualified} applicants.

This intuition--that a disparate impact analysis may look different when conducted on qualified applicants versus overall applicants--directly corresponds to the worldview that relevant construct features may differ across groups. The persistence of this assumption in the credit setting raises the question of \textit{which} differently distributed construct features courts and regulators consider to nevertheless be legitimate bases for decision-making, and which they do not. While perhaps no unbiased individual would assert that qualities such as trustworthiness differ in protected groups, they may feel that there could be cultural differences, or differences caused by structural discrimination, or differences induced by the self-selection of applicants, that are nonetheless valid bases for loan approval.

Fairness metrics based on raw outcome rates, then, should be of more concern in a discrimination case if a model relies on data that are meant to be predictive for different \textit{reasons} than traditional data are. In other words, if the predictive utility of some data is explained by its association with constructs that should \textit{not} differ across groups (such as trustworthiness), differences in raw outcomes should be less justifiable from a business necessity perspective; we discuss this situation further in Section~2.2.2.

\subsubsection{Group-level statistics on predicted and actual outcomes}

A large body of work in fair ML focuses on equalizing some statistic relating actual and predicted outcomes across different demographic subgroups. We attempt to shed insight on how each might be relevant to a fair lending case.
Some of these qualities cannot be simultaneously satisfied by any decision procedure except under specific circumstances \cite{kleinberg2016inherent}, and judging the relative importance of each metric forces us to surface our worldview assumptions and moral intuitions about when disparate outcomes are wrong.

Metrics related to \textit{sufficiency} measure the extent to which the classifier's score is equally predictive for different groups: If a model satisfies ``sufficiency," given the score, outcome is independent from the sensitive attribute. This implies further information about the sensitive attribute will not improve the model's accuracy— the score is \textit{sufficient}. \citet{fairml_book} point out that "sufficiency often comes for free (at least approximately) as a consequence of standard machine learning practices", as all available predictive information should be exploited by an optimal model. The oft-cited metric of \textit{group calibration} measures a model's closeness to a stronger condition which implies sufficiency.

Sufficiency aligns with a specific kind of moral logic. Legal scholar Deborah Hellman has argued that enforcing sufficiency would uphold the intuitive notion that "everyone is entitled to be treated by the most accurate test available (or feasible, or imaginable)" \citep{hellman}: if there were information in the data that could have helped predict an outcome, it should have been utilized. However, this interpretation of "fairness" does not align well with the disparate impact doctrine, which is triggered by the distribution of outcomes. Further, emphasizing a classifier's decision-making skills does not take into account the differing relative badness of false positives (qualified applicants denied credit) and false negatives (unqualified applicants receiving credit). In the credit setting, anti-discrimination law is much more concerned with the former.

On the other hand, \textit{separation}, otherwise known as \textit{equalized odds}, allows correlation between the score and the sensitive attribute to the extent that it is justified by the target variable. It requires that score distributions be equal between protected and unprotected individuals \textit{within} the groups of qualified and unqualified individuals. \citet{kozodoi} argue that separation is a good measure of fairness for credit because it ``accounts for the imbalanced misclassification costs of the customer, and, as these imbalanced costs also exist for the financial institution, separation is also able to consider the interests of the loan market." The relaxation of separation that qualified individuals from each group receive credit at the same rate is called \textit{equality of opportunity} \cite{hardt_eo}. This closely aligns with ECOA precedent suggesting only the acceptance rates of "qualified applicants" should be compared. \citet{hellman} argues that the \textit{ratio} of false positives and false negatives is a normatively meaningful statistic that should be equalized across groups, corresponding to a different relaxation of separation. In the credit setting, while we generally think of access to credit as a uniformly positive thing, if an applicant gets a loan they cannot pay back it is ultimately bad for them, perhaps indicating that we should balance the risk of false positives with false negatives.

Famously, sufficiency and separation cannot generally be achieved simultaneously \cite{chouldecova}. This relates to the generally accepted fact that rates of qualified and unqualified applicants (as defined by the data) may differ across groups. Essentially, if the input data contains associations with group membership, a \textit{sufficient} classifier will learn that the groups should have differing score distributions in order to be as predictive as possible--thus violating \textit{separation}.

Further, while we can certainly relate these statistics to the principles of nondiscrimination, \emph{they are only meaningful to the extent that the observed data are meaningful.} If the observed decisions were generated in a historically discriminatory or otherwise problematic manner, fairness with respect to those decisions does not imply fairness with respect to the "true" or desired variable of creditworthiness. This is why \citet{wachter_bias_preserving} call metrics in this group "bias-preserving." We discuss when this is likely to happen in the credit setting in 2.2.2.

\subsubsection{Individual fairness}

Individual fairness \citep{dwork_individual} captures the intuition that individuals who are similar with respect to the decision task should receive similar decisions. Per the framework of \citet{impossibility}, it is the construct feature space or construct decision space in which we would like to measure similarity.
Individual fairness enjoys ideological alignment with what many would consider to be fair decision-making; \citet{binns} has argued that it shares a motivation with Aristotle's conception of justice as consistency.

While individual fairness does not explicitly concern itself with protected class status, a central point in ECOA, it implicitly encodes the notion that a model should \textit{not} be sensitive to differences which are \textit{un}related to the construct features. \citet{binns} pointed out that, if "task-relatedness" is interpreted as a normative choice, individual fairness can be utilized towards egalitarian goals just as group parity metrics are. In other words, we can attempt to implement individual fairness in a way that preemptively judges protected class information to be "task-unrelated." In this sense, individual fairness can be seen as having relevance in the credit discrimination setting.

The language used to describe illegal or immoral discrimination can also be interpreted as justifying the use of an individual fairness metric to measure discrimination. The phrase "similarly situated," in particular, is a common refrain in descriptions of illegal discrimination against individuals. As legal scholar Winnie Taylor argues, "if equal credit opportunity means anything, surely it means the opportunity to be evaluated the same as other applicants similarly situated. This cuts to the essence of illegal discrimination" \citep{taylor}. To describe the contrapositive of this sentiment, \citet{Kiviat_2021} points out that in empirical work, it has been demonstrated that in economic matters, "Americans tend to define fairness through differentiation, assuming that people are different in ways that usually call for unequal allocations."

Individual fairness is also related to conditional marginal effects test, which has been used by the CFPB to analyze potentially discriminatory lending practices. In a 2015 report, they state, "The marginal effect expresses the absolute change in denial probability associated with being a member of a prohibited basis group... [the agency] also considers a conditional marginal effect, which provides the increased chances of denial for a group \textit{holding all other factors constant}, and thus controls for other, legitimate credit characteristics that may affect the probability of denial" \cite{cfpb_2015}. If membership in a sensitive group has a conditional marginal effect of 0, this means that applicants from two different groups who are identical in their input features should have a similar chance of approval--aligning with a view of individual fairness which uses input data as a similarity metric.

Importantly, however, individual fairness cannot be guaranteed based on observational criteria unless we assume that those observations are themselves unbiased \cite{impossibility}. Similarly, when the CFPB holds factors constant that are considered "legitimate credit characteristics," they are making worldview statements about the construct validity of those observational features. For this reason, efforts to use metrics related to individual fairness reasons as evidence for discrimination or nondiscrimination must rely on a deep consideration of the meaning of the data available.

\subsubsection{Causal and counterfactual reasoning}

There have been many attempts to measure illegal discrimination using causal reasoning. The term "causal inference" refers to a broad spectrum of methods and perspectives \cite{winship_and_morgan, judea_pearl}, but in essence, the goal of applying it to discrimination attempts to answer the question, "Does a protected attribute \textit{cause} a particular decision outcome?" To answer this question with causal logic, one might turn to comparing an actual outcome to a certain "what-if" scenario called a counterfactual. For instance, to determine whether a system discriminated against a black individual who was denied a loan, one might try to estimate what would have happened if the individual were white. To do this analysis, practitioners assert or discover a model of the different cause and effect relationships between relevant variables and use them to make inferences about the counterfactual scenario.

Using causal models, one can additionally attempt to distinguish between "direct" and "indirect" effects of a sensitive attribute. Some methods for learning fair models involve measuring "effects of [sensitive attributes] that are mediated by other attributes, keeping only those effects carried along paths deemed fair" \cite{hu_phenomenal}. This work posits that features like gender or race may be causally related to information that one might assert is nonetheless a valid basis for decision-making, such as GPA or department choice in the setting of graduate admissions.

Causal reasoning methods seem to closely match language used to describe discrimination in the abstract, as proponents of causal inference often point out \cite{fairml_book}. Causal logic is also often used by humans in practice for moral reasoning about decision-making in general
\cite{kiviat_moral_2019}. However, applying this line of thinking to discrimination in algorithms suffers from both conceptual and practical limitations.
On a practical level, specifying causal models requires making assumptions that cannot be validated by observational criteria and introduce complicated questions about how to understand relationships between human categories \cite{fairml_book}.
A full treatment of the conceptual critique is beyond the scope of this paper, but we very roughly summarize a line of work from Kohler-Haussman and Hu here: because discrimination is a ``thick ethical concept," which both describes and evaluates actions, it cannot be defined in terms of a causal model \cite{kohler2018eddie}. Further, traits modeled as direct or indirect "effects" of social categories are are often in fact constitutive features of those categories and relate to what makes discrimination distinctly morally problematic \cite{hu2020s}.
For these reasons, fair ML research based on measuring or improving causal and counterfactual metrics of fairness are unlikely to easily translate to enforcement or compliance with anti-discrimination law in lending.

\subsection{Specific discrimination risks in algorithmic lending}\label{sec:fairml-lending}

Machine learning algorithms complicate the interpretation of fair lending law by blurring the line of what it means for a policy to be facially neutral: even if an algorithm does not have access to protected class information, it may have been intentionally or unintentionally trained in a way that makes the ultimate policy not-so-neutral.
In this section, we use results from fair ML literature -- largely quantified in terms of the fairness metrics discussed previously -- to determine where and how bias is likely to occur in a credit modeling setting:
Firstly, if a model is trained primarily on data pertaining to a certain demographic group, that model may perform disproportionately well on that demographic group compared to others; this manifests through the problem of credit invisibility.
Secondly, if the observed features used to train the model introduce group skew from the "true" process being modeled, a model may pick up on or exacerbate these effects; these concerns are raised in a novel way by models trained on alternative data.
Thirdly, the extent to which the first two issues introduce disparities through learning are affected by other modeling choices.

\subsubsection{Sampling processes and credit invisibility}

Algorithms developed with ML techniques improve when exposed to more and more historical data. Intuitively, if training data is less available for some subpopulation of individuals, a model trained on the whole population may have performance disparities when evaluated on the groups individually. Further, since many performance metrics which are used to optimize models are constructed as averages of some kind of error-based cost across the population, these metrics are primarily driven by the model's performance on the majority class.

Both theoretical and empirical work provide evidence for the general principle that a group's under-representation in a data set can lead to group fairness disparities. \citet{chen_why} show that a learning procedure's expected performance disparities over a distribution can be additively decomposed into bias, variance, and noise components, and note that disparities caused by a difference in variances can be caused by differences in sample sizes across the groups. \citet{gendershades}'s seminal work on performance disparities in gender classification models across skin tones found that popular facial analysis benchmarking datasets are overwhelmingly white and male, and discovered that many commercial facial analysis software systems were disproportionately wrong on darker females.

This source of unfairness in machine learning is relevant in the credit setting because of the effects of \textit{credit invisibility}. In America, millions of people are "unscorable" because of their lack of credit history, and therefore face barriers to accessing credit. The CFPB recently found that income is strongly correlated with having a scored credit record, and that "Blacks and Hispanics are more likely than Whites or Asians to be credit invisible or to
have unscored credit records" \cite{cfpb_invisibles}. This matters because individuals who have never accessed credit are inherently missing from credit-report-based datasets that could be used for the supervised learning of creditworthiness.
Even if they are "scorable," individuals who have historically applied for and were rejected from loans are also by definition missing from training datasets based on the outcomes of those particular loans.
Recall that the data used to train a ML model must contain both observed features and observed decisions such as whether a historical applicant ended up defaulting on a loan. If it is not known whether an individual would have defaulted on a loan, they cannot be directly included in the supervised learning problem; therefore, we should expect low performance on those subgroups in models trained directly on historically issued loans, making them less fair from a statistical group fairness perspective.

If, however, we are concerned about the equality of \emph{outcomes} of models trained on historically issued loans, the "fairness" narrative of training on selectively labelled data may be different. \citet{harvard_garbage} suggest that under certain conditions, if a prior selection policy was biased \textit{against} a certain group, a machine learning model trained on approved applicants disproportionately \textit{favor} that group.

Of course, credit modelers know that they are missing information about applicants who were denied loans. Proceeding to only analyze the accepted applicants is called the "known good-bad" approach, but creditors usually attempt to incorporate information about the rejected applicants into their model; according to the FDIC, in the bank-issued credit card sector, "certain inferences are made to break down the rejected applicants into good and bad accounts. This procedure, known as reject inferencing, makes certain assumptions on how rejected applicants would have performed had they been accepted and attempts to mitigate any accept-only bias of the sample" \cite{fdic_ch8}. Recent work has suggested that different methods for reject inferencing may have different fairness properties \cite{coston_2021}.

There is no simple solution to correcting for the known problems induced by sampling biases. Critically, measuring (and optimizing for) the group fairness of models on the limited labelled data available using the "known good-bad" approach will produce misleading or harmful results \cite{coston_2020, kallus_zhou}. Further research studying the conditions when this hidden fairness problem arises will provide intuition with which to interpret the methodologies employed by particular lenders.

\subsubsection{Observational bias and implications for alternative data}

The nature of the specific data attributes, or features, used as inputs to a machine learning model also have an effect on the fairness of that algorithm. Recall from the framework proposed by \citet{impossibility} that an algorithm being learned by the supervised learning process is a function from the \textit{observed} feature space to the \textit{observed} decision space. The observational processes which imperfectly capture the construct features and decision can thus add group skew to the "true" relationship between construct features and construct decisions in the resulting algorithm, even when protected class attributes are not directly accessible to the algorithm through the observed features.

One way this can happen is when the observed \textit{decisions}  were generated by an explicitly discriminatory process, thus skewing the mapping from the construct decision space to the observed decision space. An example of this would be using performance reviews made by an individual with a personal bias against women as the target variable of a hiring model. If the goal of the algorithm is to make nondiscriminatory decisions, this is a poor choice of observed decision. 
Additionally, as we stated in Section~\ref{sec:fairml-defs}, if the target variable itself is biased, fairness metrics which rely on ``true" labels in the training data will be misleading.

The "observed decision" of loan repayment has generally been treated as an appropriate measurement of creditworthiness for the purposes of a business necessity defense \cite{credit_discrimination, barocas2016big}. However, if factors involved in certain instances of loan repayment are unlikely to generalize to future conditions, this may present a challenge for that argument. For instance, suppose pandemic-induced conditions disproportionately caused a certain protected group of people to default at a higher rate than others. Since the pandemic conditions may not repeat themselves in the future, the measured default variable during this time may not be relevant from a business necessity perspective, and the statistical fairness criteria cannot bolster the business necessity argument.

Another way group differences can manifest in an algorithm is if groups who are similar in the construct feature space appear different in the observed \textit{feature} space because the corresponding observational process is affected by cultural differences or discrimination. \citet{reconciling_legal} provides an example in the context of auto insurance, in which insurers would like to measure the construct feature quality of risk aversion. In this scenario, we suppose risky non-Asian drivers would choose to drive red cars more often than low-risk drivers because they are perceived as flashy or ostentatious. But it is possible that Asian drivers who drive red cars do so because red is considered a lucky color, and are no riskier than non-red-car-driving Asians. \citet{Kiviat_2021} calls such data, which "improperly conflate[s] morally distinct situations and behaviors,"  \textit{morally heterogeneous}--and finds that Americans often think using this kind of data in decision-making can be "unfair."

To determine whether observational processes are inducing bias in a credit modeling dataset, we need to take a stance on whether or not group differences can preexist in the construct feature space. Credit scores, for example, have repeatedly shown to be differently distributed across groups \cite{past_imperfect}, yet this data is seen by regulators as being related enough to creditworthiness to suffice as a basis for underwriting \textit{despite the resulting disparate outcomes}. In other words, it is implicitly being touted as a valid, low-bias measurement of a relevant construct feature, such as financial stability. \citet{impossibility} call this assumption "What you See is What You Get"--the idea that any group disparities seen in observed data are due to group disparities in construct features and are therefore an appropriate basis for decision-making.
\citet{Kiviat_2021} has shown that data which have a "logical relatedness" to a task at hand is generally seen as fair to use for that task.
Again, if evidence arises that an observational process in the training data is erroneous or generated by discriminatory processes, the claim that the data is related to creditworthiness is weaker.

For this reason it is important to vet novel, "alternative" data sources for measurement validity with respect to construct features and potential for group skew.
"Alternative data" refers to information that lenders may use for credit decisions but that is "not typically found in the consumer’s credit files," including data regarding recurring payments for utilities and rent, or cash flow data regarding deposit accounts \cite{aclu_rfi}. This strategy is gaining significant traction; \citet{Jagtiani} has uncovered evidence that online lenders are increasingly using non-traditional data to underwrite their loans. Turning to alternative data sources is meant to address the ``thin file" problem of unscored and underscored credit applicants, and in some cases this may be an appropriate solution to that problem. FinRegLab found that cash flow data provided "independent predictive value across all [demographic] groups" for credit risk and loan performance \cite{finreglab}, thus appealing to the concept of sufficiency.

However, other variables have been controversial, such as educational data in the case of Upstart. \citet{hurley_credit_2017} have reported that all kinds of data--social media profiles, technology usage, and ``how quickly a user scrolls through terms of service"--have been used for underwriting purposes by fintech companies. In general, data should receive heightened scrutiny if, as some of these features seem to be, they are attempts to measure construct features that \emph{should not vary across groups}, such as personality traits or intelligence.

\subsubsection{The importance of model complexity}

The degree to which the sampling and modeling problems described above actually affect a machine learning algorithm depends on the chosen model class and training procedure. In particular, it relates to a model type's \textit{capacity}, which measures how well it can capture complex patterns in data. For example, new, "powerful" ML tools like gradient boosting and deep learning are high-capacity, whereas traditional linear models are low-capacity. These differences are salient in a fairness sense: Low-capacity models on data which is disparately predictive between classes may result in low cost-based fairness. On the other hand, high-capacity models on predictive data can be have more unequal outcomes than simple models if there is bias in the labels.

Low capacity models on disparately predictive data can discriminate in a "statistical group fairness" sense. \citet{chen_why}'s decomposition of statistical group fairness measures shows that differences in a model's bias can cause group fairness to deteriorate, which happens when "the chosen model class is not flexible enough to fit both groups well." For this reason, a sufficiently complex model trained on culturally diverse data could be "less" biased than a simple one (for instance, able to capture the different meanings of Red in the car example). In theory, this could present a problem for banks, who traditionally use simpler models such as logistic regression \cite{frb_study, barocas2016big} if they apply these models to alternative, "morally heterogeneous" data as in the car example. Interestingly, however, studies have show that advanced modeling techniques to more \textit{traditional} data does not necessarily improve outcomes \cite{noise}, so low model capacity may not be acting as a source of bias in this context.

On the other hand, high-capacity models on very biased but predictive historical data can amplify discrimination more than a low-capacity model can. For instance, a sufficiently complex model trained on  biased hiring data could be more biased than a simple one, by being able to more precisely pick up on gender through resume items using combinations of words instead of single words. This presents a risk in the case of new, more cutting-edge ``fintech" companies which are more likely to be experimenting with high-capacity models such as gradient boosting and random forests. \citet{band_DI} has shown that outcome rate disparities of a model are mathematically connected to how predictive the input data are of a protected attribute, and there is also evidence applying data mining to HMDA \textit{is more predictive of race than zip code} \cite{gillis}. For these reasons, the accidental encoding of racial information in a high-capacity model trained on biased data is a real danger in the credit setting.

\section{Regulatory opportunities to enforce fairness in machine learning}\label{sec:regulatory-opps}

This paper has demonstrated that ML fairness research suggests that machine learning and alternative data present fair lending risks that should be of concern to regulators. Now we tackle the implications for the enforcement of fair lending regulation. In particular, we present two broad strategies that regulators could pursue to identify and mitigate the fairness risks identified in Section 2.

\subsection{Expanding the collection and analysis of protected class attributes}

Protected class information on loan applicants is necessary to effectively measure and mitigate unfairness, which as Section 2 has argued, is a real threat--yet it is still legally risky to collect. When the CFPB was first established, \citet{taylor} suggested they were well-positioned to remove the general ban preventing creditors to collect protected class information; in theory, they can directly amend or change Regulation B. In practice this would be politically difficult, for reasons discussed in Section 1.1.2. However, the CFPB has made steps to increase the amount of data available in this space by changing the requirements in the context of small business loans. Further action incentivizing or requiring the collection of protected class data would enable interventions for detecting and preventing discrimination as well as expanding access to credit.

The first benefit of this data is unrelated to machine learning specifically: protected class information on applicants for loans would enable external oversight of lenders as HMDA data does in the mortgage setting. As Regulation C states, a major purpose of the HMDA data requirement is to ``assist in identifying potential discriminatory lending patterns and enforcing antidiscrimination statutes;'' as mentioned previously, many have argued that a similar provision would be helpful for the same tasks \cite{gao, taylor}. The CFPB would no longer have to rely on BISG or related methods to impute sensitive data for their audits.

The collection of protected class attributes in credit data also expands the range of tools for developing fair models available to algorithm developers. Many of the quantitative fairness frameworks discussed in Section 2 lend themselves to an optimization problem: training or modifying a model to be fair(er). Some of these tools offer interventions to modify the training dataset, the way the model is trained, or tweaking the model after it has been learned in the usual way \cite{comparison}. Access to protected class information in the training dataset is generally required to implement most of these methods, although workarounds have been proposed \cite{jiahao_unawareness}.

The act of directly forcing an algorithm to conform to fairness metrics raises potential legal issues. Methods that require access to a sensitive attribute of an applicant at decision time are unlikely to gain traction in the credit setting, since ECOA specifically prohibits using the protected class of an individual in a credit decision. This includes options such as training different models for different protected classes, which may be legal in other contexts \cite{hellman}.

What remains less clear is whether interventions which have access to \emph{historical} protected class information at \emph{training} time could be legal or even encouraged.
Some scholars have expressed concern that preemptively modifying an algorithm for fairness could be considered disparate treatment \cite{barocas2016big}, or analogous to racial quotas \cite{reconciling_legal}. However, \citet{kim}'s detailed treatment of the issue describes many situations in which race-conscious decision-making is not considered disparate treatment under anti-discrimination law. She concludes that techniques "more accurately understood as removing bias from processes," such as efforts to correct biased input data or formulate a fair problem specification, are legally permissible, whereas methods that more closely resemble a quota system will likely trigger close legal scrutiny. The qualitative differences between specific bias mitigation methods are therefore important to describe and evaluate, but there appears to be a legal path forward for regulators to encourage some of these strategies. Guidance from agencies on this issue is currently sorely lacking, and must be addressed.

There are other considerations at hand here, as modifying models to fit fairness criteria in practice can introduce other, non-legal problems. \citet{comparison} benchmarked several against a variety of fairness metrics on existing datasets, and found that they tended to be brittle and sensitive to fluctuations in dataset composition, highlighting the importance of careful experimental design when drawing conclusions about fairness. Studies employing economic methods have also shown that the long-term effects of enforcing fairness have implications for social welfare \cite{liu_and_hardt, hu_and_chen}; this relates to a broader discussion of balancing nondiscrimination with economic efficiency that is outside the scope of this paper.

Even if regulators or lenders are uncomfortable with making their traditional credit risk models fairness-aware, ECOA specifically allows special purpose credit programs to be targeted at expanding access to credit to traditionally underserved populations. Unfortunately, few lenders have taken advantage of this allowance. Protected class data could provide insight into how to effectively underwrite credit to those populations.

\subsection{Managing discrimination risk as model risk}

Several organizations \cite{brookings} have suggested that one way regulators can use their authority to mitigate discrimination risk is by treating it as any other kind of model risk and applying the relevant guidelines and standards to the development of models in the financial space. For instance, to apply SR 11-7 to discrimination risk, some suggest ``the Agencies should ensure that financial institutions have appropriate Compliance Management Systems that effectively identify and control risks related to AI systems, including the risk of discriminatory or inequitable outcomes for consumers'' \cite{aclu_rfi}.

Scholarly analyses of discrimination support the idea that developers should be held liable for unintentional algorithmic discrimination.
In his theoretical treatment of algorithmic discrimination, legal scholar Tal Zarsky considers the negligent or reckless usage of biased data to be a form of intentional implicit discrimination, in which ``firms' failure to act and prevent discrimination" is a form of intent. For this reason, he argues that "such behavior should be actively countered" and that "additional policy discussions must establish the proper standard of care this normative justification calls for on behalf of the scorers" \cite{zarsky}.

SR 11-7 emphasizes that model risk guidance emphasizes that "risk assessment should be conducted by independent actors within the institution or a third party." If financial institutions are pressured to do more self-testing using gathered or approximated protected class data, and follow this guidance, the team who developed a model may be informed that their models have undesirable fairness metrics. They may be able to use this information to develop a less discriminatory alternative without directly using protected class data. For instance, simply changing a model's overall acceptance threshold can influence the fairness statistics of a model \cite{aclu_rfi}. This "indirect" optimization may be more legally defensible than the direct optimization strategies discussed previously.

The OCC's model risk guidance also recommends continuous monitoring of models in deployment. Monitoring models for correctness is important, but developers can also monitor their models for changes in the incoming data that could affect the fairness dynamics described above. For instance, macroeconomic changes may affect changes in the underlying demographic composition of applicants, which will in turn affect observable fairness characteristics.

Specific federal guidance on how to responsibly manage bias risk could be developed based on the fair ML results discussed in this paper. For instance, the agencies could recommend that bias risk be considered and estimated when a developer chooses how to conduct reject inferencing, as discussed in Section 2. NIST has already taken steps towards developing a framework to mitigate bias risk in general; the Agencies could build off of their work or develop a credit-specific framework in parallel.

\section*{Conclusion}

Fair machine learning research has shown time and again that there is a direct relationship between a developer's practices and the fairness of those outcomes. This paper has outlined why policymakers should be concerned with fairness in credit algorithms by pointing out the specific discrimination risks that should be mitigated. Policymakers and ML researchers must work together to determine how to motivate developers to deploy fair models, as well as curate the tools and the data to do so. To do this, fair ML researchers must understand the goals of regulators in this space. Conversely, policymakers will benefit from understanding relevant results from fair ML. We view this paper in part as a call to arms for the development of a shared understanding between the two communities.

\clearpage
\markboth{References}{References}
\balance
\bibliographystyle{ACM-Reference-Format}
\bibliography{bib}

\end{document}